\documentclass[final]{arxiv-v2}

\usepackage{times}
\usepackage{epsfig}
\usepackage{graphicx}
\usepackage{amsmath}
\usepackage{amssymb}

\usepackage{amsfonts}
\usepackage{bbm}
\usepackage{booktabs}
\usepackage{algorithm}
\usepackage{algorithmic}
\usepackage{subfigure}
\usepackage{multirow}
\usepackage{footmisc}

\usepackage[pagebackref=false,breaklinks=true,colorlinks,bookmarks=false]{hyperref}

\def\cvprPaperID{4210} 
\def\confYear{CVPR 2021}

\begin{document}

\title{Bi-GCN: Binary Graph Convolutional Network}

\author{
  Junfu Wang\textsuperscript{\rm 1,2},
  Yunhong Wang\textsuperscript{\rm 2},
  Zhen Yang \textsuperscript{\rm 1,2},
  Liang Yang \textsuperscript{\rm 3},
  Yuanfang Guo \textsuperscript{\rm 1,2}\thanks{Corresponding author.}\\
    \textsuperscript{\rm 1} State Key Laboratory of Software Development Environment, Beihang University, China\\
    \textsuperscript{\rm 2} School of Computer Science and Engineering, Beihang University, China\\
    \textsuperscript{\rm 3} School of Artificial Intelligence, Hebei University of Technology, China \\
{\tt\small \{wangjunfu,yhwang,yangzhen7,andyguo\}@buaa.edu.cn,yangliang@vip.qq.com}
}


\maketitle

\pagestyle{empty}  
\thispagestyle{empty} 

\begin{abstract}
  Graph Neural Networks (GNNs) have achieved tremendous success in graph representation learning.
  Unfortunately, current GNNs usually rely on loading the entire attributed graph into network for processing.
  This implicit assumption may not be satisfied with limited memory resources, especially when the attributed graph is large.
  In this paper, we pioneer to propose a Binary Graph Convolutional Network (Bi-GCN), which binarizes both the network parameters and input node features.
  Besides, the original matrix multiplications are revised to binary operations for accelerations.
  According to the theoretical analysis, our Bi-GCN can reduce the memory consumption by an average of $\thicksim$30x for both the network parameters and input data, and accelerate the inference speed by an average of $\thicksim$47x, on the citation networks.
  Meanwhile, we also design a new gradient approximation based back-propagation method to train our Bi-GCN well.
  Extensive experiments have demonstrated that our Bi-GCN can give a comparable performance compared to the full-precision baselines. 
  Besides, our binarization approach can be easily applied to other GNNs, which has been verified in the experiments.
\end{abstract}

\section{Introduction}

  In the past few years, Graph Neural Networks (GNNs), which can learn effective representations from irregular data, have given excellent performances in various graph-based tasks \cite{gcn,gat,togcn,gin}.
  Considering the superior representation abilities of these newly developed GNNs, researchers have also applied them to many tasks, including natural language processing \cite{Text-Classification-gcn}, computer vision \cite{few-shot-gcn}, etc.
  
  Unfortunately, the current success of GNNs is attributed to an implicit assumption that the input of GNNs contains the entire attributed graph.
  If the entire graph is too large to be fed into GNNs due to limited memory resources, in both the training and inference process, which is highly likely when the scale of the graph increases, the performances of GNNs may degrade drastically.
  
  To tackle this problem, an intuitive solution is sampling, e.g., sampling a subgraph with a suitable size to be entirely loaded into GNNs.
  The sampling based methods can be classified into two categories, neighbor sampling \cite{graphsage,stgcn} and graph sampling \cite{fastgcn, cluster-gcn, graphsaint}.
  Neighbor sampling selects a fixed number of neighbors for each node in the next layer to ensure that every node can be sampled.
  Thus, it can be utilized in both the training and inference process.
  Unfortunately, when the number of layers increases, the problem of \textit{neighbor explosion} \cite{graphsaint} arises, such that both the training and inference time will increase exponentially.
  Different from neighbor sampling, graph sampling samples a set of subgraphs in the training process, which can avoid the problem of \textit{neighbor explosion}.
  However, it cannot guarantee that every node can be at least sampled once in the whole training/inference process.
  Thus it is only feasible for the training process, because the testing process usually requires GNNs to process each node in the graph.

  \begin{figure}[t]
    \centering
    \subfigure[] { 
      \label{fig1:a}
      \includegraphics[width=0.45\columnwidth]{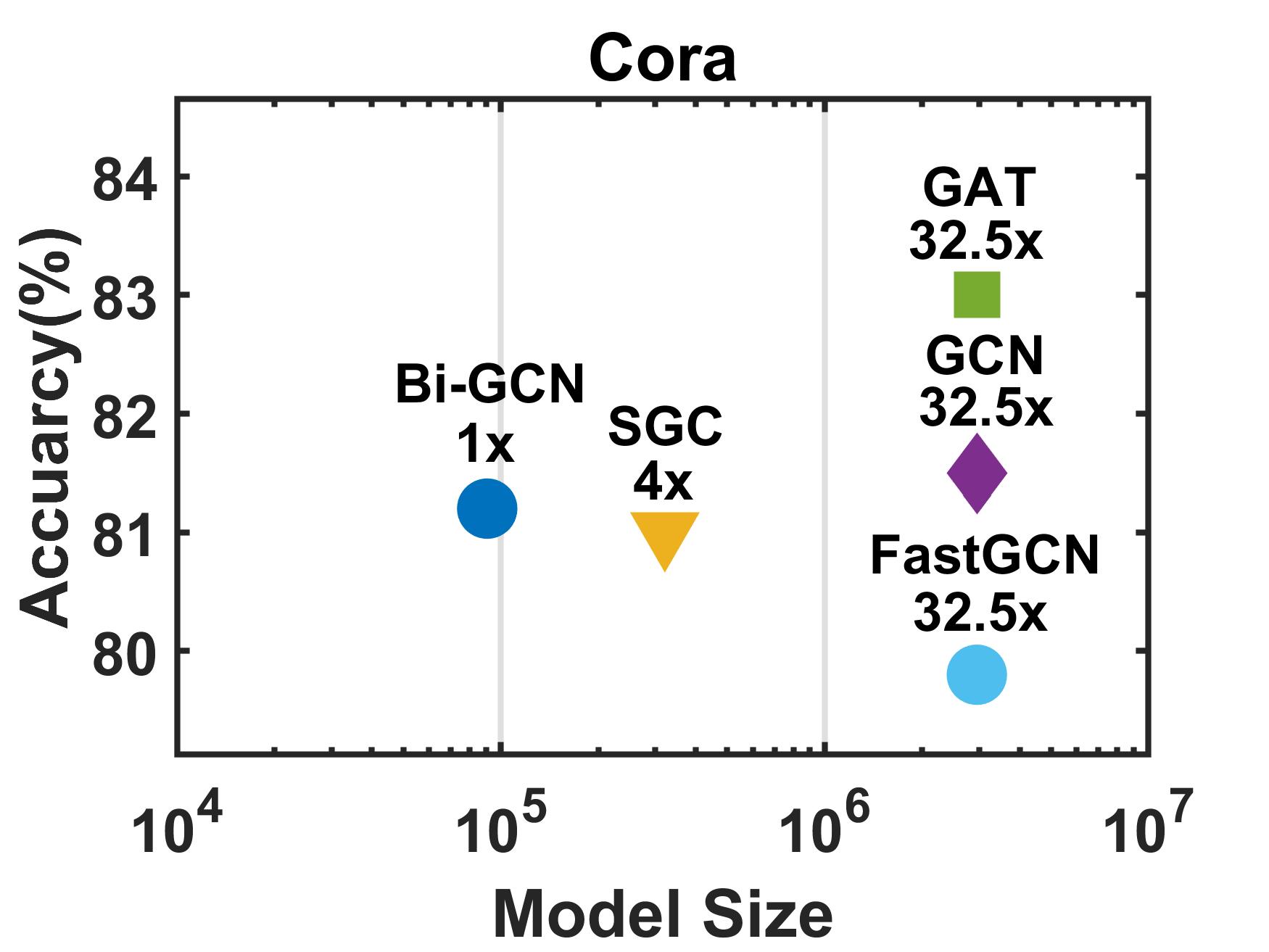}
    }
    \subfigure[] { 
      \label{fig1:b}
      \includegraphics[width=0.45\columnwidth]{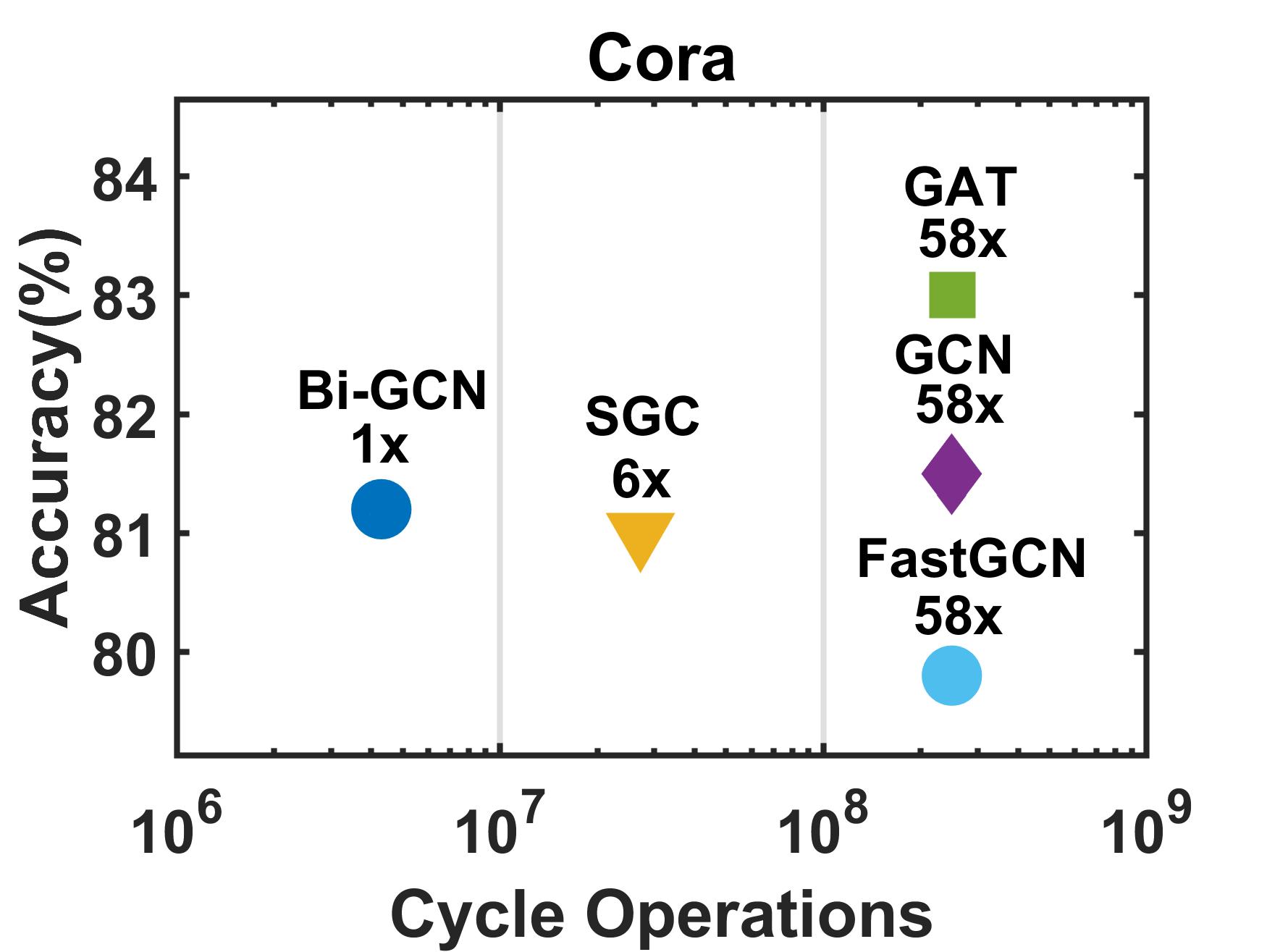}
    }
    \vspace{-0.05cm}
    \caption{Performances on the Cora dataset.
    Note that the model size is measured in bits and the number of cycle operations, which will be introduced in Sec. 5, is employed to reflect the inference speed.
    Bi-GCN gives the fastest inference speed and the lowest memory consumption with comparable accuracy.
    }
    \label{fig1}
    \vspace{-0.4cm}
  \end{figure}

  Another feasible solution is compressing the size of the input graph data and the GNN model to better utilize the limited memory and computational resources.
  Several approaches have been proposed to compress the convolutional neural networks (CNNs), such as designing shallow networks \cite{shallownetworks}, pruning \cite{compress2}, designing compact layers \cite{inception} and quantizing the parameters \cite{Binarized-Neural-Networks}.
  In quantization-based methods, binarization \cite{Binarized-Neural-Networks,xnornet,bi-real} has achieved a great success in many CNN-based practical vision tasks when a faster speed and a lower memory consumption is desired.

  However, compared to the CNN compression methods, the compression of GNNs possesses unique challenges.
  Firstly, since the input graph data is usually much larger than the GNN models, the compression of the loaded data demands more attention.
  Secondly, the GNNs are usually shallow, e.g., the standard GCN \cite{gcn} only has 2 layers, which contain less redundancies, thus the compression will be more difficult to be achieved.
  At last, the nodes tend to be similar to its neighbors in the high-level semantic space, while they tend to be different in the low-level feature space, which is different from the grid-like data, such as images, videos, etc.
  This characteristic requires the compressed GNNs to possess sufficient parameters for representations. In general, the tradeoff between the compression ratio and accuracy in the compressed GNNs requires careful designs.
  
  To tackle the memory and complexity issues, SGC \cite{sgc}, which is a 1-layered GNN, compresses GCN \cite{gcn} by removing nonlinearities and collapsing weight matrices between consecutive layers.
  This shallow GNN can accelerate both the training and inference processes with comparable performance.
  Although SGC compresses the network parameters, it does not compress the loaded data, which is the major memory consumption when processing the graphs with GNNs.

  In this paper, to alleviate the memory and complexity issue, we pioneer to propose a binarized GCN, named Binary Graph Convolutional Network (Bi-GCN), which is a simple yet efficient approximation of GCN \cite{gcn}, by binarizing the parameters and node attribute representations.
  Specifically, the binarization of the weights is performed by splitting them into multiple feature selectors and maintaining a scalar per selector to further reduce the quantization errors.
  Similarly, the binarization of the node features can be carried out by splitting the node features and assigning an attention weight to each node.
  By employing those additional scalars, more efficient information can be learned and retained efficiently.
  After binarizing the weights and node features, the computational complexity and the memory consumptions induced by the network parameters and input data can be largely reduced.
  Since the existing binary back propagation method \cite{xnornet} has not considered the relationships among the binary weights, we also design a new back propagation method by tackling this issue.
  An intuitive comparison between our Bi-GCN and the baseline methods is shown in Figure \ref{fig1}, which demonstrates that our Bi-GCN can achieve the fastest inference speed and lowest memory consumption with a comparable accuracy compared to the standard full-precision GNNs.

  Our proposed Bi-GCN can reduce the redundancies in the node representations while maintain the principle information.
  When the number of layers increases, Bi-GCN also gives a more obvious reductions of the memory consumptions of the parameters and effectively alleviates the overfitting problem.
  Besides, our binarization approach can be easily applied to other GNNs.

  The contributions are summarized as follows:
  \begin{itemize}
  \item We pioneer to propose a binarized GCN, named Binary Graph Convolutional Network (Bi-GCN), which can significantly reduce the memory consumptions by $\thicksim$30x for both the network parameters and input node attributes, and accelerate the inference by an average of $\thicksim$47x, on the citation networks, theoretically.
  \item We design a new back propagation method to effectively train our Bi-GCN, by considering the relationships among the binary weights in the back propagation process.
  \item With respect to the significant memory reductions and accelerations, our Bi-GCN can also give a comparable performance compared to the standard GCN on four benchmark datasets.
  \end{itemize}

  \section{Related Work}
  \subsection{Sampling Based GNNs}
  Sampling is an effective method that allows GNNs to process larger graphs with limited memory.
  Current sampling methods can be categorized into two categories, neighbor sampling \cite{graphsage,stgcn} and graph sampling \cite{fastgcn, cluster-gcn, graphsaint}.
  GraphSAGE \cite{graphsage} gives an empirical number of the sampled neighbors and extends the GNNs to inductive learning.
  VRGCN \cite{stgcn} reduces the sampling size by maintaining the embedding of each node from the previous iteration, which requires a doubled memory consumption.
  Meanwhile, FastGCN \cite{fastgcn} samples a subgraph in each layer to accelerate the training process, which sacrifices the classification accuracy.
  ClusterGCN \cite{cluster-gcn} groups the nodes by graph clusting methods, which demands additional complexity for the clustering.
  GraphSAINT \cite{graphsaint} proposes an edge sampling method with low variance and apply GCN \cite{gcn} to the sampled subgraphs.
  Besides, DropEdge \cite{dropedge} generates the subgraphs randomly and DropConnection \cite{dropconnection} adaptively samples the subgraphs, to alleviate the overfitting problem.
  
  \subsection{CNN Binarization Methods}
  Convolutional Neural Networks (CNNs) suffer from certain issues, such as high computational costs and etc.
  Binarization, as a promising type of techniques in network compression, has been widely utilized to reduce the memory and computation costs for CNNs.
  BinaryConnect \cite{BinaryConnect} binarizes the network parameters and replaces most of the floating-point multiplications with floating-point additions.
  Binarynet \cite{Binarynet} further binarizes the activation function and uses the XNOR (not-exclusive-OR) operations to accelerate the inference process.
  XNOR-Net \cite{xnornet} proposes a scalar based binarization approach and successfully applies it to the popular CNNs, such as ResNet \cite{resnet} and GoogLeNet \cite{googlenet}.
  
  \section{Preliminaries}
  \subsection{Notations}
  Here, we define the notations utilized throughout this paper.
  We denote an undirected attributed graph as $\mathcal{G}=\left\{\mathcal{V},\mathcal{E},X\right\}$ with the vertex set $\mathcal{V} =\{v_i\}_{i=1}^N$ and edge set $\mathcal{E}=\{e_i\}_{i=1}^E$. 
  Each node $v_i$ contains a feature $X_i\in\mathbb{R}^d$.
  $X\in\mathbb{R}^{N\times d}$ is the collection of all the features in all the nodes. 
  $A=[a_{ij}] \in\mathbb{R}^{N\times N}$ is the adjacency matrix which reveals the relationships between each pair of vertices, i.e., the topology information of $\mathcal{G}$.
  $d_i=\sum_j a_{ij}$ stands for the degree of node $v_i$ and
  $D=diag(d_1, d_2, \dots, d_n)$ represents the degree matrix corresponding to the adjacency matrix $A$. 
  Then, $\hat{A} = A + I$ is the adjacency matrix of the original topology with self-loops and $\hat{D}$ is its corresponding degree matrix with $\hat{D}_{ii} = \sum_j\hat{a}_{ij}$.
  Note that we employ the superscript "$(l)$" to represent the $l$-th layer, e.g., $H^{(l)}$ is the input node features to the $l$-th layer.
  
  \subsection{Graph Convolutional Network}
  Graph Convolutional Network (GCN) \cite{gcn} has become the most popular graph neural network in the past few years.
  Since our binarization approach takes GCN as the basis GNN, we give a brief review of GCN here.
  
  Given an undirected graph $\mathcal{G}$, the graph convolution operation can be described as
  \begin{equation}
    H^{(l+1)} = \sigma(\tilde{A}H^{(l)}W^{(l)}),
    \label{gcn-propagation}
  \end{equation}
  where $\tilde{A}=\hat{D}^{-\frac{1}{2}}\hat{A}\hat{D}^{-\frac{1}{2}}$ is a sparse matrix, and $W^{(l)}\in\mathbb{R}^{d_{in}^{(l)}\times d_{out}^{(l)}}$ contains the learnable parameters.
  Note that $H^{(l+1)}$ is the output of the $l$-th layer and the input of the $(l+1)$-th layer, and $H^{(0)} = X$.
  $\sigma$ is the non-linear activation function, e.g., ReLU.

  From the perspective of spatial methods, the graph convolution layer in GCN can be decomposed into two steps, where $\tilde{A}H^{(l)}$ is the aggregation step and $H^{(l)}W^{(l)}$ is the feature extraction step.
  The aggregation step tends to constrain the node attributes in the local neighborhood to be similar.
  After that, the feature extraction step can easily extract the commonalities between the neighboring nodes.
  
  GCN typically utilizes a task-dependent loss function, e.g., the cross-entropy loss for the node classification tasks, which is defined as
  \begin{equation}
    \mathcal{L} = -\sum_{v_i\in\mathcal{V}^{label}}\sum_{c=1}^{C}Y_{i,c}log(\tilde{Y}_{i,c}),
    \label{gcn-loss}
  \end{equation}
  where $\mathcal{V}^{label}$ stands for the set of the labelled nodes, $C$ denotes the number of classes, $Y$ represents the ground truth labels, and $\tilde{Y}=softmax(H^{(L)})$ are the predictions of the $L$-layered GCN.

  \section{Binary Graph Convolutional Network}
  In this section, we propose our Binary Graph Convolution Network (Bi-GCN), a binarized version of the standard GCN. 
  As mentioned in previous section, a graph convolution layer can be decomposed into two steps, aggregation and feature extraction.
  In Bi-GCN, we only focus on binarizing the feature extraction step, because the aggregation step possesses no learnable parameters (which yields negligible memory consumption) and it only requires a few calculations (which can be neglected compared to the feature extraction step).
  Therefore, the aggregation step of the original GCN is maintained.
  For the feature extraction step, we binarize both the network parameters and node features to reduce the memory consumptions.
  To reduce the computational complexities and accelerate the inference process, the XNOR (not-exclusive-OR) and bit count operations are utilized, instead of the traditional floating-point multiplications.
  Finally, we design an effective back-propagation algorithm for training our binarized graph convolution layer.

  \begin{figure*}[t]
    \centering
    \includegraphics[width=1.65\columnwidth]{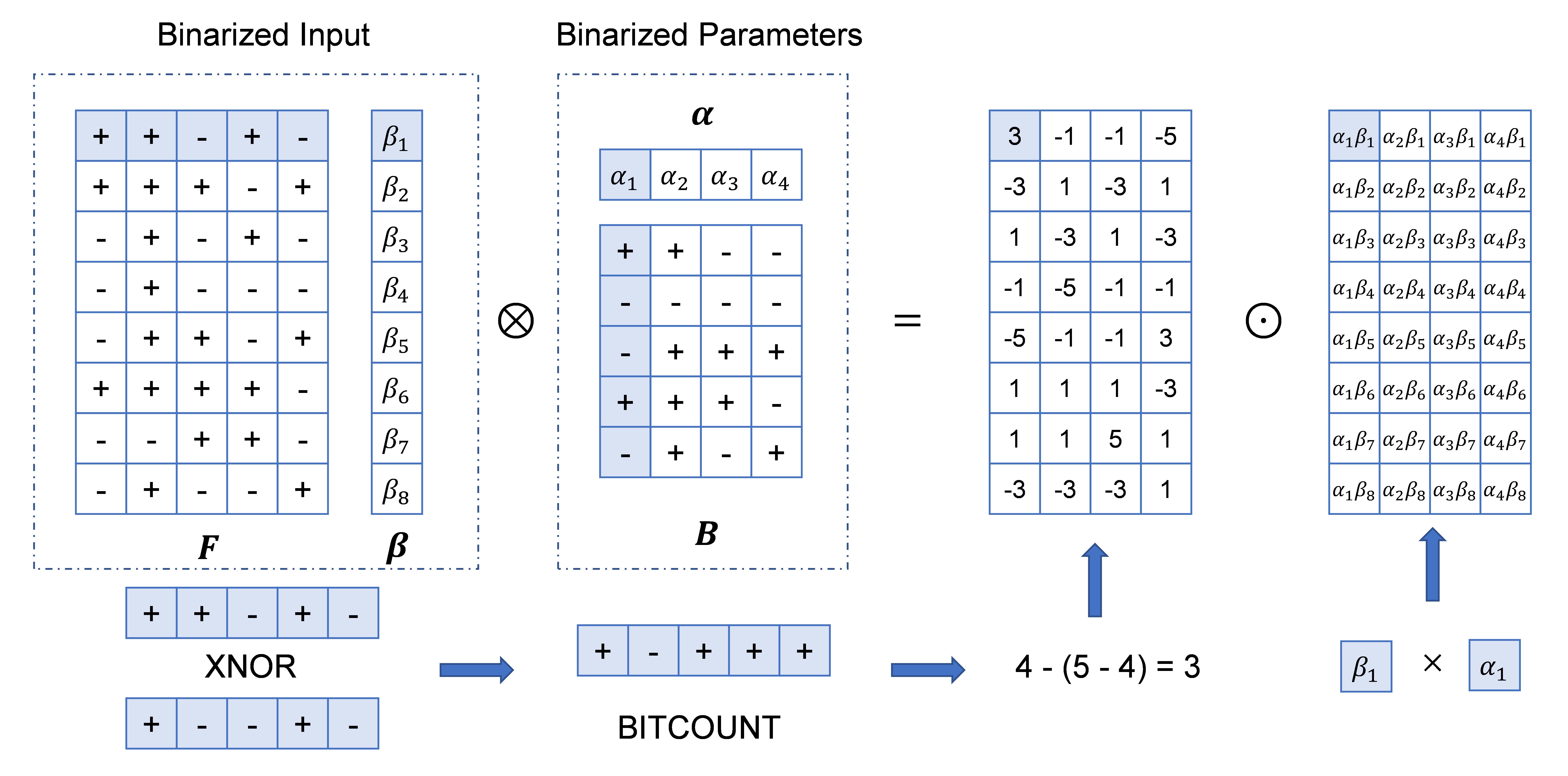}
    \caption{An example of binary feature extraction step.
    Both the input features and parameters will be binarized to binary matrices.
    $\otimes$ denotes the binary matrix multiplication defined in Sec. 4 and $\odot$ represents the element-wise multiplication.
    }
    \label{fig2}
    \vspace{-0.2cm}
  \end{figure*}

  \subsection{Binarization of the Feature Extraction Step}
  
  Based on the vector binarization algorithm \cite{xnornet}, we can perform the binarization to the feature extraction step $Z^{(l)}=H^{(l)}W^{(l)}$ in the graph convolution shown in Eq. \ref{gcn-propagation}.
  Note that for this feature extraction (matrix multiplication), we adopt the bucketing \cite{qsgd} method to generalize the binary inner product operation to the binary matrix multiplication operation.
  Specifically, we split the matrix into multiple buckets of consecutive values with a fixed size and perform the scaling operation separately.
  
  \subsubsection{Binarization of the Parameters}
  Since each column of the parameter matrix of the $l$-th layer $W^{(l)}$ serves as a feature selector in the computation of $Z^{(l)}$, each column of $W^{(l)}$ is splitted as a bucket, i.e., a vector.
  Let $\alpha^{(l)}=(\alpha_1^{(l)}, \alpha_2^{(l)}, ..., \alpha_{d_{out}^{(l)}}^{(l)})$, which are the scalars for each bucket.
  Let $B^{(l)}=(B_1^{(l)},B_2^{(l)},...,B_{d_{out}^{(l)}}^{(l)})\in\{-1, 1\}^{d_{in}^{(l)}\times d_{out}^{(l)}}$ be the binarized buckets of $W^{(l)}$.
  Then, based on the vector binarization algorithm, the optimal $B^{(l)}$ and $\alpha^{(l)}$ can be easily calculated by
  \begin{equation}
    B_j^{(l)}=sign(W^{(l)}_{:,j}),
    \label{B}
  \end{equation}
  \begin{equation}
    \alpha_j^{(l)}=\frac{1}{d_{out}^{(l)}}||W^{(l)}_{:,j}||_1,
    \label{alpha}
  \end{equation}
  where $W^{(l)}_{:,j}$ represents the $j$-th column of $W^{(l)}$. 
  It can be approximated via
  \begin{equation}
    W^{(l)}_{:,j}\approx\tilde{W}^{(l)}_{:,j}=\alpha_j^{(l)} B^{(l)}_j.
    \label{app-W}
  \end{equation}
  Based on Eq. \ref{app-W}, the graph convolution operation with binarized weights can then be described as
  \begin{equation}
    H^{(l+1)} \approx H^{(l+1)}_{p} = \sigma(\tilde{A}H^{(l)}\tilde{W}^{(l)}),
  \end{equation}
  where $H^{(l+1)}_{p}$ is the binary approximation of $H^{(l+1)}$ with the binarized parameters $\tilde{W}^{(l)}$.
  The binarization of the parameters can reduce the memory consumption by a factor of $\thicksim$30x, compared to the parameters with full precision, which will be proven in Sec. 5.

  \subsubsection{Binarization of the Node Features}
  Due to the over-smoothing issue \cite{deeperinsights} induced by the current graph convolution operation, current GNNs are usually shallow, e.g., the vanilla GCN only contains 2 graph convolution layers.
  Although the future GNNs may possess a larger model, the data sizes of commonly employed attributed graphs are usually much larger than the current model size.
  To reduce the memory consumption of the input data, which is mostly induced by the node features, we also perform binarization to the node features which will be processed by the graph convolutional layers.

  To binarize the node features, we split $H^{(l)}$ into row buckets based on the constraints of the matrix multiplication to compute $Z^{(l)}$, i.e., each row of $H^{(l)}$ will conduct an inner product with each column of $W^{(l)}$.
  Let $\beta^{(l)}=(\beta_1^{(l)}, \beta_2^{(l)}, ..., \beta_{N}^{(l)})$ denote the scalars for each bucket in $H^{(l)}$.
  Let $F^{(l)}=(F_1^{(l)}; F_2^{(l)}; ...; F_N^{(l)})\in\{-1, 1\}^{N\times d_{in}^{(l)}}$ be the binarized buckets.
  Then, with the vector binarization algorithm, the optimal $\beta$ and $F$ can be computed by

  \begin{equation}
    \beta_i^{(l)}=\frac{1}{N}||H_{i,:}^{(l)}||_1,
    \label{beta}
  \end{equation}
  \begin{equation}
    F_i^{(l)}=sign(H^{(l)}_{i,:}),
    \label{F}
  \end{equation}
  where $H_{i,:}^{(l)}$ represents the $i$-th row of $H^{(l)}$. 
  Then, the binary approximation of $H^{(l)}$ can be obtained via
  \begin{equation}
    H^{(l)}_{i,:}\approx\tilde{H}^{(l)}_{i,:}=\beta_i^{(l)} F^{(l)}_i.
  \end{equation}
  Intuitively, $\beta$ can be considered as the node-weights for the feature representations.
  At last, the graph convolution operation with binarized weights and node features can be formulated as
  \begin{equation}
    H^{(l+1)} \approx H^{(l+1)}_{ip} = \tilde{A}\tilde{H}^{(l)}\tilde{W}^{(l)}.
    \label{H-ip}
  \end{equation}

  Note that this binarization of the node features, i.e., the input of the graph convolutional layer, also possesses the ability of activation, thus we do not employ specific activation functions (such as ReLU).
  Similar to the binarization of the weights, the memory consumption of the loaded attributed graph data can be reduced by a factor of $\thicksim$30x compared to the vanilla GCN.
  
  \subsubsection{Binary Operations} 
  With the binarized graph convolutional layers, we can accelerate the calculations by employing the XNOR and bit-count operations instead of the floating-point additions and multiplications. 
  Let $\zeta^{(l)}$ represent the approximation of $Z^{(l)}$. Then,
  \begin{equation}
    Z_{ij}^{(l)}\approx\zeta_{ij}^{(l)} =\beta_i^{(l)}\alpha_j^{(l)} F_{i,:}^{(l)}\cdot B_{:,j}^{(l)}.
    \label{cdot-op}
  \end{equation}
  Since each element of $F^{(l)}$ and $B^{(l)}$ is either -1 or 1, the inner product between these two binary vectors can be replaced by the binary operations, i.e., XNOR and bit count operations.
  Then, Eq. \ref{cdot-op} can be re-written as
  \begin{equation}
    \zeta_{ij}^{(l)} =\beta_i^{(l)}\alpha_j^{(l)} F_{i,:}^{(l)}\circledast B_{:,j}^{(l)},
    \label{bi-op}
  \end{equation}
  where $\circledast$ denotes a binary multiplication operation using the XNOR and a bit count operations.
  The detailed process is illustrated in Figure \ref{fig2}.
  Therefore, the graph convolution operation in the vanilla GCN can be approximated by
  \begin{equation}
    H^{(l+1)} \approx H^{(l+1)}_{b} = \tilde{A}\zeta^{(l)},
  \end{equation}
  where $\zeta^{(l)}$ is calculated via Eq. \ref{bi-op} and $H^{(l+1)}_{b}$ is the final output of the $l$-th layer with the binarized parameters and inputs.
  By employing this binary multiplication operation, the original floating point calculations can be replaced with identical number of  binary operations and a few extra floating-point calculations.
  It will significantly accelerate the processing speed of the graph convolutional layers.

  \begin{algorithm}[t]
    \caption{Back propagation process for training a binarized graph convolutional layer}
    \label{alg1}
    \renewcommand{\algorithmicrequire}{\textbf{Input:}}
	  \renewcommand{\algorithmicensure}{\textbf{Output:}}
    \begin{algorithmic}[1]
    \REQUIRE Gradient of the layer above $\frac{\partial{\mathcal{L}}}{\partial{H^{(l+1)}}}$
    \ENSURE Gradient of the current layer $\frac{\partial{\mathcal{L}}}{\partial{H^{(l)}}}$
    \STATE Calculate the gradients of $\tilde{W}^{(l)}$ and $\tilde{H}^{(l)}$
    
    {$\frac{\partial{\mathcal{L}}}{\partial{\zeta^{(l)}}}=\tilde{A}^T\cdot\frac{\partial{\mathcal{L}}}{\partial{H^{(l+1)}}}$}

    {$\frac{\partial{\mathcal{L}}}{\partial{\tilde{W}^{(l)}}}=(\tilde{H}^{(l)})^T\cdot\frac{\partial{\mathcal{L}}}{\partial{\zeta^{(l)}}}$}
    
    {$\frac{\partial{\mathcal{L}}}{\partial{\tilde{H}^{(l)}}} = \frac{\partial{\mathcal{L}}}{\partial{\zeta^{(l)}}}\cdot\tilde{W}^{(l)}$}
    
    \STATE Calculate $\frac{\partial{\mathcal{L}}}{\partial{H^{(l)}}}$ via Eq. \ref{bp-H}
    \STATE Calculate $\frac{\partial{\mathcal{L}}}{\partial{W^{(l)}}}$ via Eq. \ref{bp-W}
    \STATE Update $\tilde{W}^{(l)}$ with the gradient $\frac{\partial{\mathcal{L}}}{\partial{W^{(l)}}}$
    \STATE \textbf{return} $\frac{\partial{\mathcal{L}}}{\partial{H^{(l)}}}$
    \end{algorithmic}
  \end{algorithm}

  \subsection{Binary Gradient Approximation Based Back Propagation}
  The key parts of our training process include the choice of the loss function and the back-propagation method for training the binarized graph convolutional layer.
  The loss function employed in our Bi-GCN is the same as the vanilla GCN, as shown in Eq. \ref{gcn-loss}.
  Since the existing back propagation method \cite{xnornet} has not considered the relationships among the binary weights, to perform the back-propagation for the binarized graph convolutional layer, the gradient calculation is desired to be newly designed.
  
  To calculate the actual propagated gradient for the $l$-th layer, the binary approximated gradient $\frac{\partial{\mathcal{L}}}{\partial{\tilde{H}^{(l)}}}$ is employed to approximate the gradient of the original one as \cite{Binarized-Neural-Networks,xnornet},
  \begin{equation}
    \frac{\partial{\mathcal{L}}}{\partial{H^{(l)}}} \approx\frac{\partial{\mathcal{L}}}{\partial{\tilde{H}^{(l)}}}\mathbbm{1}_{|\frac{\partial{\mathcal{L}}}{\partial{\tilde{H}^{(l)}}}|<1}.
    \label{bp-H}
  \end{equation}
  Note that $\mathbbm{1}_{|r|<1}$ is the indicator function, whose value is 1 when $|r|<1$, and vice versa.
  This indicator function serves as a hard tanh function which preserves the gradient information.
  If the absolute value of the gradients becomes too large, the performance will be degraded.
  Thus, the indicator function also serves to kill certain gradients whose absolute value becomes too large.
  
  The gradient of network parameters is computed via another gradient calculation approach.
  Here, a full-precision gradient is employed to preserve more gradient information.
  If the gradient of the binarized weights $\frac{\partial{\mathcal{L}}}{\partial{\tilde{W}^{(l)}}}$ is obtained, $\frac{\partial{\mathcal{L}}}{\partial{W_{ij}^{(l)}}}$ can then be calculated as
  \begin{equation}
    \begin{aligned}
    \frac{\partial{\mathcal{L}}}{\partial{W_{ij}^{(l)}}} =& \frac{\partial{\mathcal{L}}}{\partial{\tilde{W}_{:,j}}^{(l)}}\cdot\frac{\partial{\tilde{W}_{:,j}^{(l)}}}{\partial{W_{ij}^{(l)}}}\\
    =&\frac{1}{d_{in}^{(l)}}{B_{ij}^{(l)}}\sum_k\frac{\partial{\mathcal{L}}}{\partial{\tilde{W}_{kj}^{(l)}}}\cdot{B_{kj}^{(l)}}+\alpha_j^{(l)}\cdot\frac{\partial{\mathcal{L}}}{\partial{\tilde{W}_{ij}^{(l)}}}\cdot\frac{\partial{B_{ij}^{(l)}}}{\partial{W_{ij}^{(l)}}}.
    \end{aligned}
    \label{bp-W}
  \end{equation}

  To compute the gradient for the sign function $sign(\cdot)$, the straight-through estimator (STE) function \cite{st-estimating} is employed, where $\frac{\partial{sign(r)}}{\partial{r}} = \mathbbm{1}_{|r|<1}$. 
  The back-propagation process is summarized in Algorithm \ref{alg1}.

  \section{Analysis}
  In this section, we theoretically analyze the performance of our Bi-GCN, i.e., the compression ratio of the model size and the loaded data size, as well as the acceleration ratio, respectively, compared to the full-precision (32-bit floating-point representation) GCN.
  \subsection{Model Size Compression}
  Let the parameters of each layer in the full-precision GCN be denoted as $W^{(l)}\in\mathbb{R}^{d_{in}^{(l)}\times d_{out}^{(l)}}$, which contains $(d_{in}^{(l)}\times d_{out}^{(l)})$ floating-point parameters.
  On the contrary, the $l$-th layer in our Bi-GCN only contains $(d_{in}^{(l)}\times d_{out}^{(l)})$ binary parameters and $d_{out}^{(l)}$ floating-point parameters.
  Therefore, the size of the parameters can be reduced by a factor of 
  \begin{equation}
    PC^{(l)} =\frac{32d_{in}^{(l)}d_{out}^{(l)}}{d_{in}^{(l)}d_{out}^{(l)}+32d_{out}^{(l)}} = \frac{32d_{in}^{(l)}}{d_{in}^{(l)}+32}.  
    \label{pc-l}
  \end{equation}
  
  According to Eq. \ref{pc-l}, the compression ratio of the parameters for the $l$-th layer is depending on the dimension of input node features. 
  For example, a 2-layered Bi-GCN, whose hidden layer contains 64 neurons, can achieve a $\thicksim$31x model size compression ratio compared to the full-precision GCN on Cora dataset. 
  Although the memory consumption of the network parameters is smaller than the input data for the vanilla GCN, our binarization approach still contributes.
  Currently, many efforts have already been made to construct deeper GNNs  \cite{deepgcn,dropedge,fastanddeepgcn}.
  As the number of layers increases, the reductions on the memory consumptions will become much larger and this contribution will become more significant.
  


  \subsection{Data Size Compression}
  Currently, the loaded data tends to contribute the majority of the memory consumptions.
  In the commonly employed datasets, the node features tends to contribute the majority of the loaded data.
  Thus, a binarization of the loaded node features can largely reduce the memory consumptions when GNNs process the datasets.
  Note that the data size of the node features is employed as an approximation of the entire loaded data size in this paper, because the edges in commonly processed attribute graph is usually sparse and the size of the division mask is also small.

  Let the loaded node features be denoted as $X\in\mathbb{R}^{N\times d}$, where $N$ is the number of nodes and $d$ is the number of features per node.
  Then, the full-precision $X$ contains $N\times d$ floating-point values.
  In our Bi-GCN, the loaded data $X$ can be binarized, and $N\times d$ binary values and $N$ floating-point values can be obtained.
  Thus, the size of the loaded data $X$ can be reduced by a factor of
  \begin{equation}
    DC =\frac{32Nd}{Nd+32N} = \frac{32d}{d+32}.
    \label{DC}
  \end{equation}
  
  According to Eq. \ref{DC}, the compression ratio of the loaded data size is depending on the dimension of the node features.
  In practical, Bi-GCN can achieve an average reduction of memory consumption with a factor of $\thicksim$30x, which indicates that a much bigger attributed graph can be entirely loaded with identical memory consumption.
  For some inductive datasets, we can then successfully load the entire graph or use a bigger sub-graph than that in the full-precision GCN.
  The results of data size compression can be found in Tables \ref{table-trans} and \ref{table-ind}.

  \subsection{Acceleration}
  After the analysis of memory consumptions, the analysis of acceleration of our Bi-GCN, compared to GCN, is performed.
  Let the input matrix and the parameters of the $l$-th layer possess the dimensions $N\times d_{in}^{(l)}$ and $d_{in}^{(l)}\times d_{out}^{(l)}$, respectively.
  The original feature extraction step in GCN requires $Nd_{in}^{(l)}d_{out}^{(l)}$ addition and $Nd_{in}^{(l)}d_{out}^{(l)}$ multiplication operations. 
  On the contrary, the binarized feature extraction step in our Bi-GCN only requires $Nd_{in}^{(l)}d_{out}^{(l)}$ binary operations and $2Nd_{out}$ floating-point multiplication operations. 
  According to \cite{xnornet}, the processing time of performing one cycle operation, which contains one multiplication and one addition, can be utilized to perform 64 binary operations.
  Then, the acceleration ratio for the feature extraction step of the $l$-th layer can be calculated as
  \begin{equation}
    S^{(l)}_{fe} =\frac{Nd_{in}^{(l)}d_{out}^{(l)}}{\frac{1}{64}Nd_{in}^{(l)}d_{out}^{(l)}+2Nd_{out}^{(l)}} = \frac{64d_{in}^{(l)}}{d_{in}^{(l)}+128}.  
    \label{S-fe}
  \end{equation}
  As can be observed from Eq. \ref{S-fe}, the dimension of the node features $d_{in}^{(l)}$ determines the acceleration efficiency for the feature extraction step. 
  
  For the aggregation step, the sparse matrix multiplication contains $|\mathcal{E}|d_{out}^{(l)}$ floating-point addition and $|\mathcal{E}|d_{out}^{(l)}$ floating-point multiplication operations.
  If we let the average degree of the nodes be $\overline{deg}$, then $|\mathcal{E}| = N\overline{deg}/2$.
  
  Then, the complete acceleration ratio of the $l$-th graph convolutional layer can be approximately computed via
  \begin{equation}
    \begin{aligned}
      S^{(l)}_{full} &=\frac{Nd_{in}^{(l)}d_{out}^{(l)}+|\mathcal{E}|d_{out}^{(l)}}{\frac{1}{64}Nd_{in}^{(l)}d_{out}^{(l)}+2Nd_{out}^{(l)}+|\mathcal{E}|d_{out}^{(l)}} \\
      & = \frac{64d_{in}^{(l)}+32\overline{deg}}{d_{in}^{(l)}+128+32\overline{deg}}.
    \end{aligned}
    \label{S-full}
  \end{equation}
  Note that the average degree $\overline{deg}$ is usually small in the benchmark datasets, e.g., $\overline{deg}\approx 2.0$ in the Cora dataset.
  When processing a graph with a low average node degree, the computational cost for the aggregation step, i.e., $32\overline{deg}$, usually possesses negligible effect on the acceleration ratio.
  Thus, the acceleration ratio of the $l$-th layer can be approximately computed via
  \begin{equation}
      S^{(l)}_{full} \approx S^{(l)}_{fe}.
      \label{S-app}
  \end{equation}
  Therefore, when $\overline{deg}$ is small, the acceleration ratio mainly depends on the input dimension of the binarized graph convolutional layers, according to Eqs. \ref{S-fe} and \ref{S-app}.
  The input dimension of the first graph convolutional layer equals to the dimension of the node features in the input graph.
  The input dimensions of the other graph convolutional layers equal to the dimensions of the hidden layers.
  Since the dimension of the input node features is usually large, the acceleration ratio tends to be high for the first layer, e.g., $\thicksim$59x on the Cora dataset.
  In general, the layer with a larger input dimension tends to require more calculations and can thus save more calculations with our binarization.
  For example, the acceleration ratio of a 2-layered Bi-GCN on the Cora dataset can achieve $\thicksim$59x acceleration ratio for the first layer and $\thicksim$21x for the second layer.
  In total, our 2-layered Bi-GCN can achieve $\thicksim$53x acceleration ratio on the Cora dataset.

  \begin{table}[t]
    \small
    \centering
    \caption{Datasets}
    \begin{tabular}{ccccc}
      \toprule
      Dataset & Nodes & Edges & Classes & Features\\
      \midrule
      Cora&  2,708 & 5,429 & 7 & 1,433\\
      PubMed& 19,711 & 44,338 & 3 & 500\\
      \midrule
      Flickr & 89,250 & 899,756 & 7 & 500\\
      Reddit & 232,965 & 11,606,919 & 41 & 602\\
      \bottomrule
    \end{tabular}
    \label{table-datasets}
  \end{table}

  \section{Evaluations}
  In this section, we evaluate the proposed binarization approach and our Bi-GCN on benchmark datasets for the node classification task\footnote{More experiments can be found in the supplimentary material.}.
  Note that the memory consumptions and the number of cycle operations are ideally estimated based on the specific settings of the methods and datasets.
  
  \subsection{Datasets}
  We conduct our experiments on four commonly employed datasets. 
  For the transductive learning task, two commonly utilized citation networks, i.e., Cora and PubMed \cite{citation}, which are also employed by GCN \cite{gcn}, are utilized. 
  We adopt the same data division strategy as \cite{planetoid}.
  For the inductive learning task, Flickr and Reddit are employed.
  We adopt the same data division strategy as GraphSAINT \cite{planetoid} for Flickr and GraphSAGE \cite{graphsage} for Reddit.
  The datasets are summarized in Table \ref{table-datasets}.
  %

  \begin{table*}[t]
    \small
    \centering
    \caption{Transductive learning results.(M.S., D.S., and C.O. are the abbreviations of Model Size, Data Size and Cycle Operations.)}
    \begin{tabular}{c|cccc|cccc}
      \toprule
      \multirow{2}{*}{Networks} & \multicolumn{4}{c|}{Cora} & \multicolumn{4}{c}{PubMed}\\
      \cline{2-9}
      & Accuracy & M.S. & D.S. & C.O. & Accuracy &  M.S. & D.S. & C.O.\\
      \midrule
      GCN        & \textbf{81.4 ± 0.4} & 360K            & 14.8M          & 2.50e8          &  79.0 ± 0.3 & 125.75K       & 37.6M          & 6.38e8\\
      Bi-GCN(binarize features only)& 81.1 ± 0.4 & 360K            & \textbf{0.47M} & 2.50e8          &  \textbf{79.4 ± 1.0} & 125.75K       & \textbf{1.25M} & 6.38e8\\
      Bi-GCN(binarize weights only)& 78.3 ± 1.5 & \textbf{11.53K} & 14.8M          & 2.50e8          &  75.5 ± 1.4 & \textbf{4.19K}& 37.6M          & 6.38e8\\
      Bi-GCN     & 81.2 ± 0.8 & \textbf{11.53K} & \textbf{0.47M} & \textbf{4.67e6} &  78.2 ± 1.0 & \textbf{4.19K}& \textbf{1.25M} & \textbf{1.55e7}\\
      \midrule
      GAT        & 83.0 ± 0.7 & 360.55K         & 14.8M          & 2.51e8          &  79.0 ± 0.3 &  126.27K      & 37.6M          & 6.44e8\\
      FastGCN    & 79.8 ± 0.3 & 360K            & 14.8M          & 2.50e8          &  79.1 ± 0.2 & 125.75K       & 37.6M          & 6.38e8\\
      SGC        & 81.0 ± 0.0 & 39.18K          & 14.8M          & 2.72e7          &  78.9 ± 0.0 & 5.86K         & 37.6M          & 2.98e7\\
      \bottomrule
    \end{tabular}
    \label{table-trans}
  \end{table*}

  \begin{figure}[t]
    \centering
    \subfigure[] { 
      \label{fig4:a}
      \includegraphics[width=0.46\columnwidth]{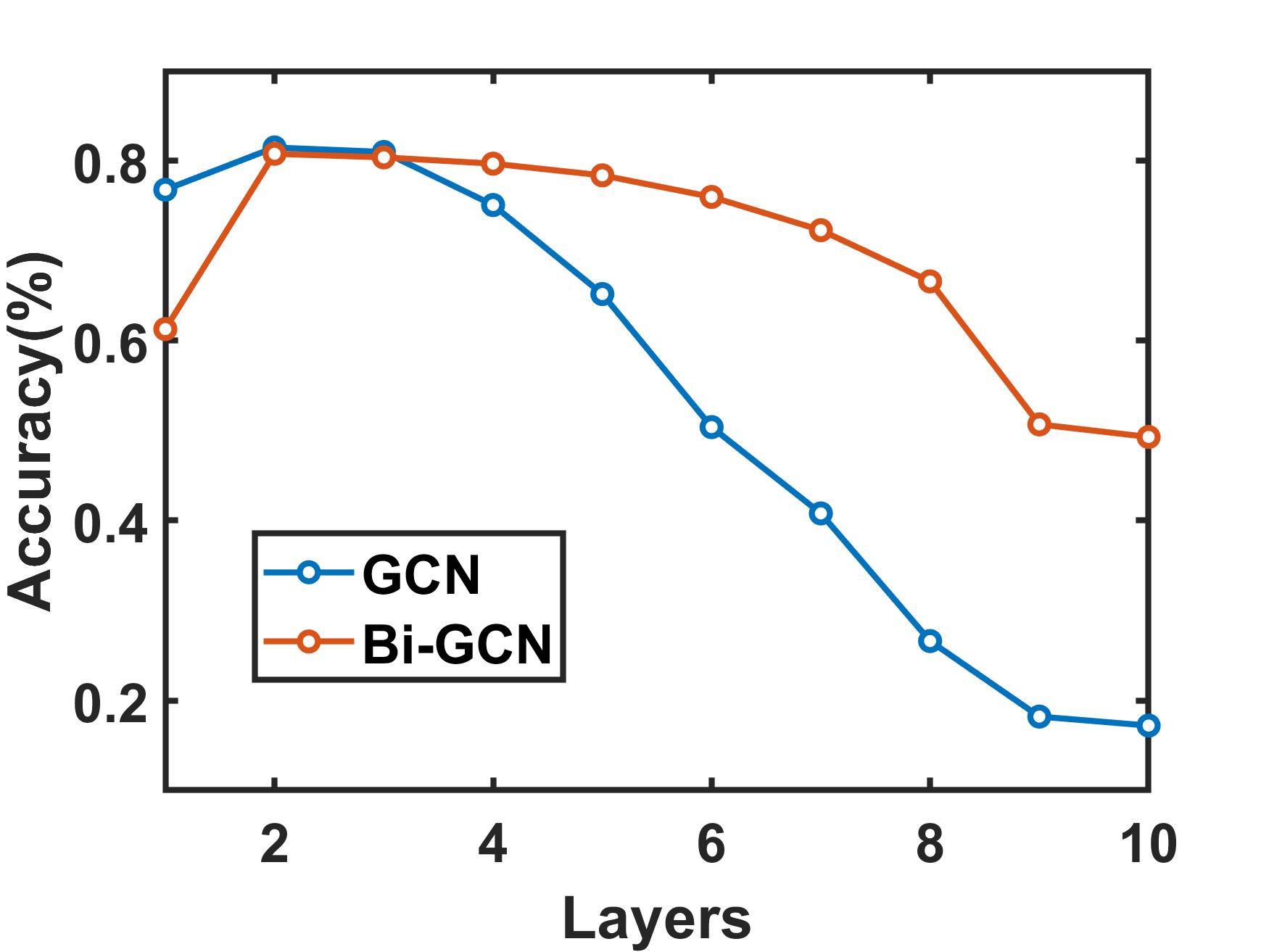}
    }
    \subfigure[] { 
      \label{fig4:b}
      \includegraphics[width=0.44\columnwidth]{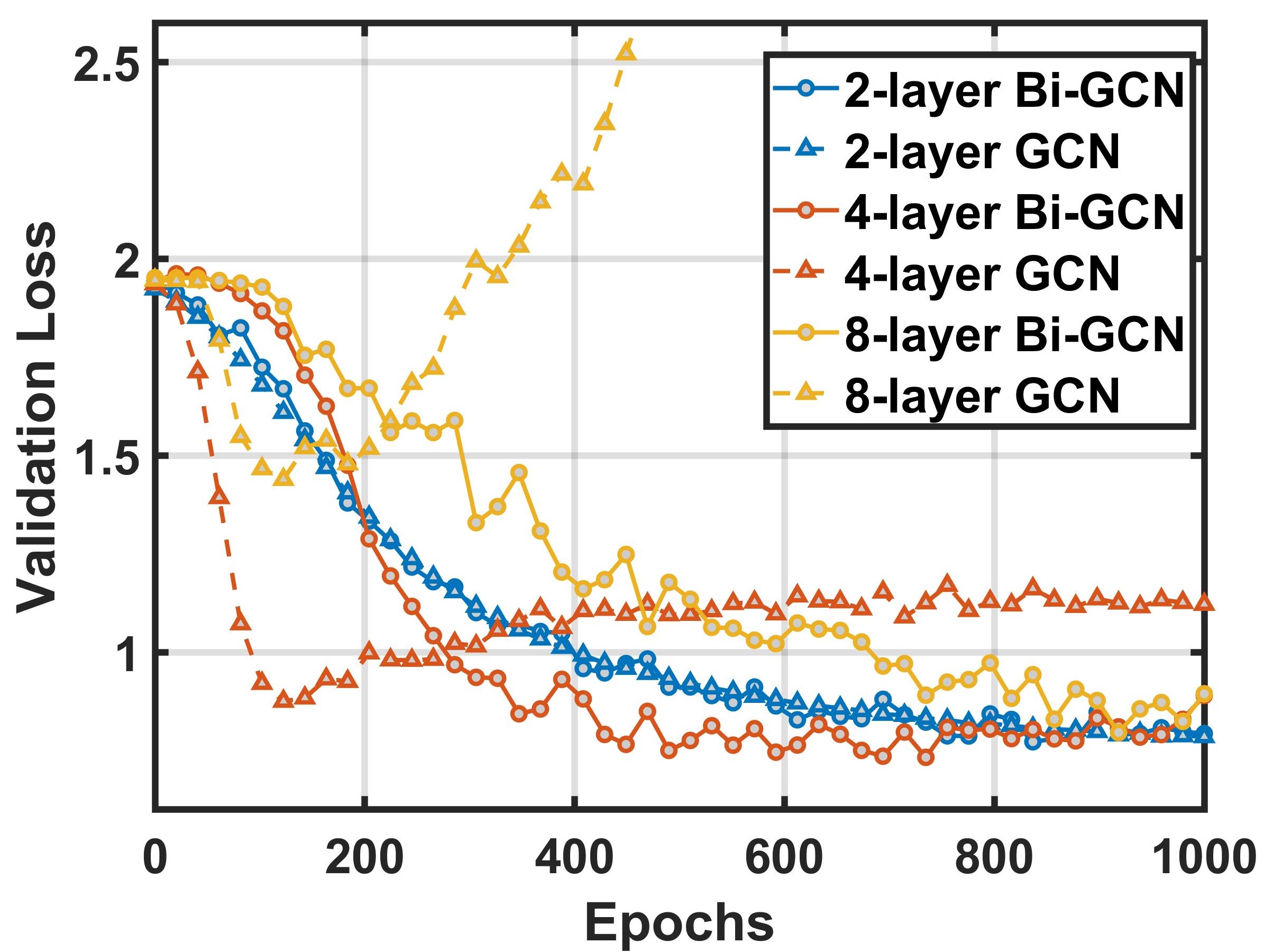}
    }
    \vspace{-0.1cm}
    \caption{Comparisons of accuracy and validation loss on Cora.}
    \label{fig4}
    \vspace{-0.4cm}
  \end{figure}

  \begin{figure}[t]
    \centering
    \subfigure[] { 
      \label{fig5:a}
      \includegraphics[width=0.45\columnwidth]{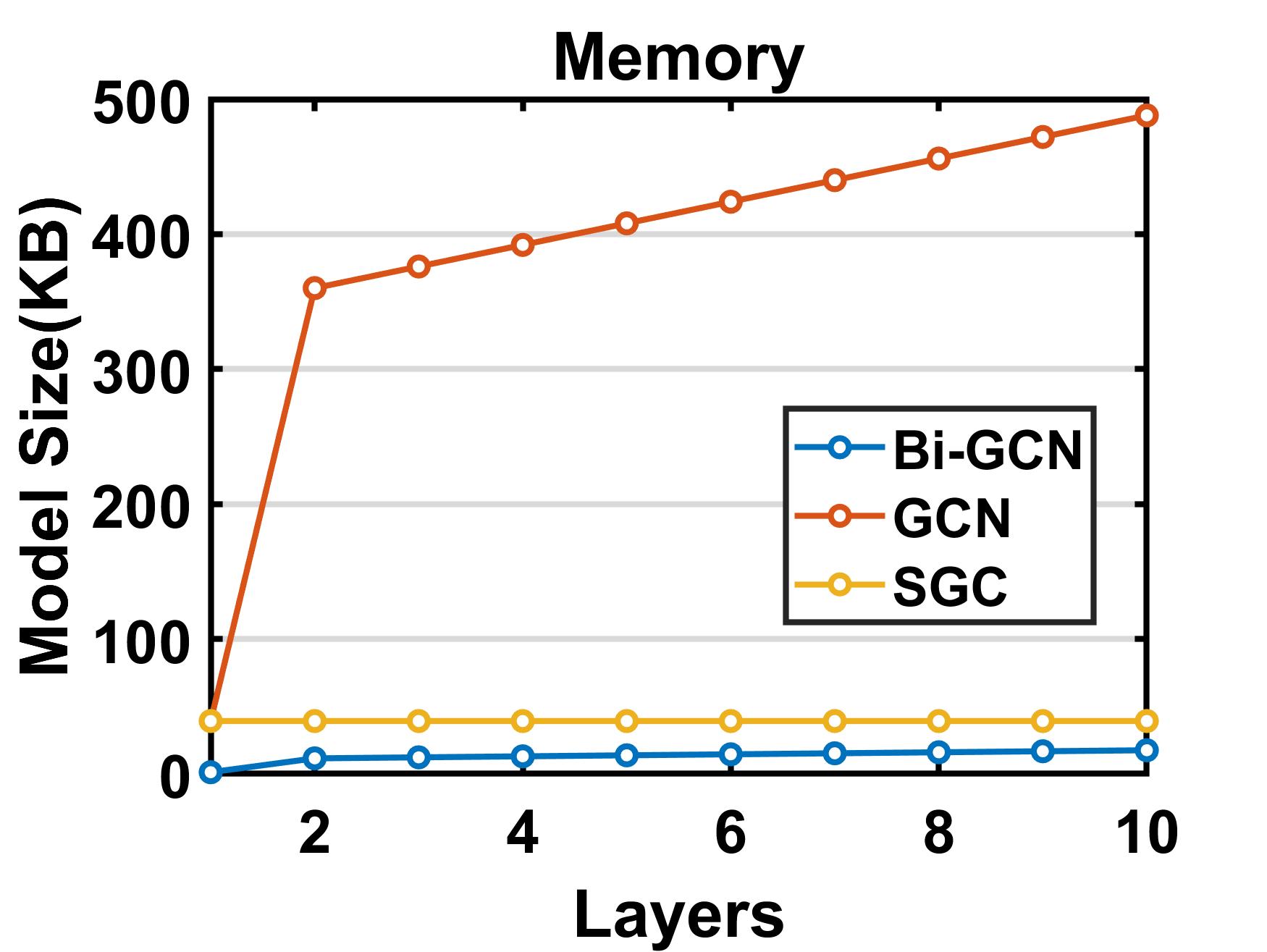}
    }
    \subfigure[] { 
      \label{fig5:b}
      \includegraphics[width=0.45\columnwidth]{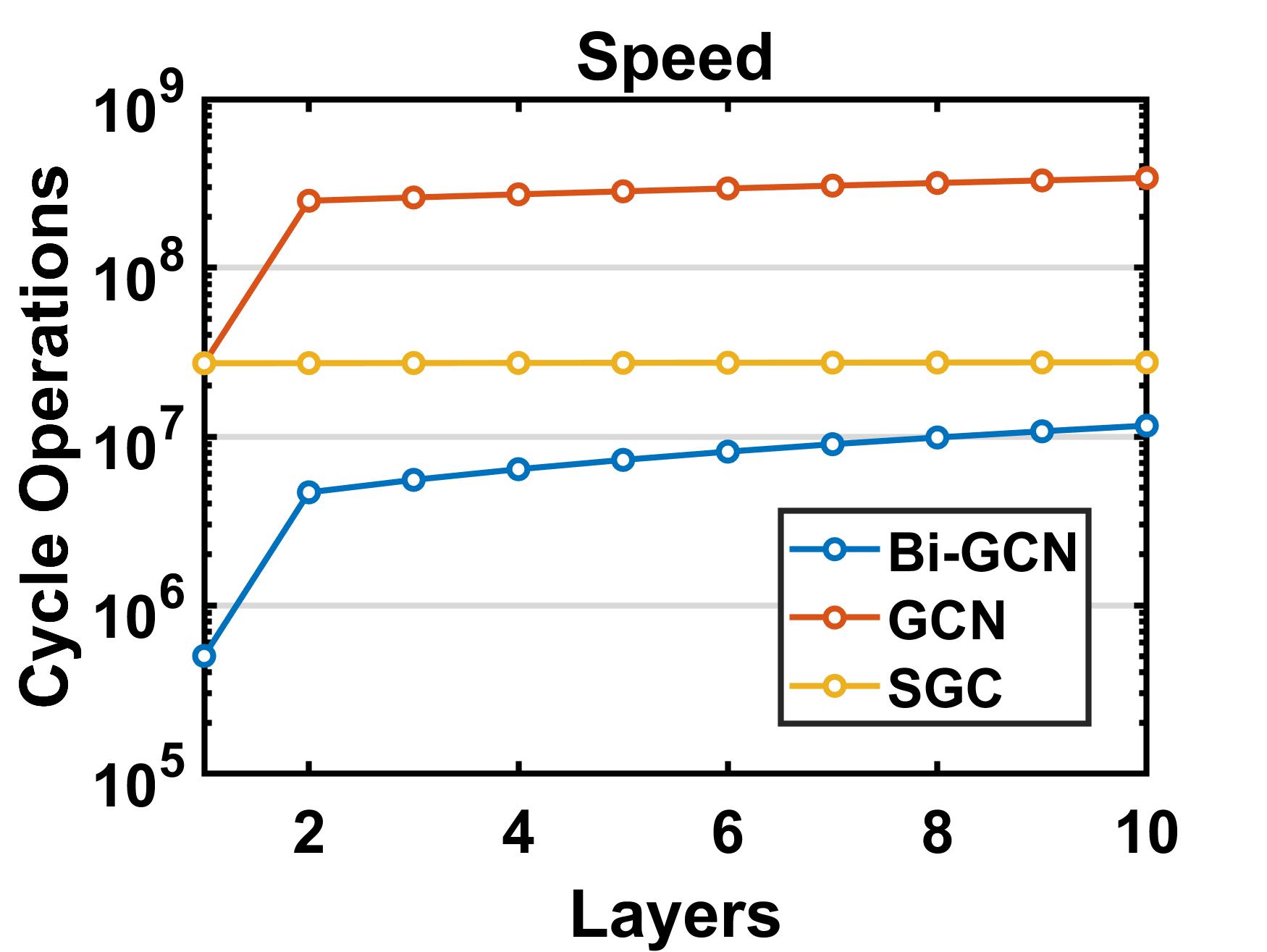}
    }
    \vspace{-0.1cm}
    \caption{Comparisons of memory consumption and inference speed on Cora.
    }
    \label{fig5}
    \vspace{-0.5cm}
  \end{figure}

  \subsection{Setups}
  For the transductive learning task, we select a 2-layered GCN \cite{gcn} with 64 neurons in the hidden layer as the baseline.
  Our Bi-GCN is obtained by binarizing this GCN.
  The evaluation protocol in \cite{gcn} is applied.
  In the training process, GCN and Bi-GCN are both trained for a maximum of 1000 epochs with an early stopping condition at 100 epochs, by using the Adam \cite{adam} optimizer with a learning rate of 0.001.
  The dropout layers are utilized in the training process with a dropout rate of 0.4, after binarizing the input of the intermediate layer.
  We initialize the full-precision weights by Xavier initialization \cite{glorot}.
  A standard batch normalization \cite{bn} (with zero mean and variance being one) is applied to the input feature vectors in Bi-GCN. 
  Note that we also investigate the influences of different model depths on classification performance. 
  All the hyperparameters are set to be identical to the 2-layered case.
  
  
  For the inductive learning task, we select the inductive GCN \cite{graphsage}, GraphSAGE \cite{graphsage} and GraphSAINT \cite{graphsaint} as our baselines. 
  Note that a 2-layered GraphSAINT model is employed for fair comparisons.
  The settings from their own literatures are employed.
  We will binarize all the feature extraction steps to generalize their corresponding binarized version.
  The hyper-parameters in our binarized models are set to be identical to their full-precision version.

  \begin{table*}[t]
    \small
    \centering
    \caption{Inductive learning results. (M.S., D.S., and C.O. are the abbreviations of Model Size, Data Size and Cycle Operations.)}
    \begin{tabular}{c|cccc|cccc}
      \toprule
      \multirow{2}{*}{Networks} & \multicolumn{4}{c|}{Reddit} & \multicolumn{4}{c}{Flickr}\\
      \cline{2-9}
      & F1-micro & M.S. & D.S. & C.O. & F1-micro & M.S. & D.S. & C.O.\\
      \midrule
      inductiveGCN & \textbf{93.8 ± 0.1} &643.00K& 534.99M & 4.18e10 &\textbf{50.9 ± 0.3} &507.00K&170.23M & 1.18e10\\
      Bi-inductiveGCN & 93.1 ± 0.2 & \textbf{21.25K}  & \textbf{17.61M}    & \textbf{4.18e9}& 50.2 ± 0.4&\textbf{16.87K}&\textbf{5.66M}&\textbf{4.65e8} \\
      \midrule
      GraphSAGE& 95.2 ± 0.1 & 1286.00K & 534.99M & 8.01e10 & \textbf{50.9 ± 1.0}&1014.00K &170.23M& 2.34e10 \\
      Bi-GraphSAGE & \textbf{95.3 ± 0.1} & \textbf{42.51K} & \textbf{17.61M} & \textbf{4.92e9} & 50.2 ± 0.4&\textbf{33.74K}& \textbf{5.66M}& \textbf{6.93e8}\\
      \midrule
      GraphSAINT& \textbf{95.9 ± 0.1} & 1798.00K& 534.99M  & 1.13e11&\textbf{52.1 ± 0.1}&1526.00K&170.23M&3.53e10\\
      Bi-GraphSAINT& 95.7 ± 0.1 & \textbf{139.62K} &\textbf{17.61M} & \textbf{1.04e10} & 50.8 ± 0.2&\textbf{65.25K}&\textbf{5.66M}&\textbf{1.28e9}\\
      \bottomrule
    \end{tabular}
    \label{table-ind}
  \end{table*}


 
  \subsection{Results}
  \subsubsection{Comparisons}
  The results of the transductive learning tasks are shown in Table \ref{table-trans}. 
  As can be observed, our Bi-GCN gives a comparable performance compared to the full-precision GCN and other baselines.
  Meanwhile, our Bi-GCN can achieve an average of $\thicksim$47x faster inference speed and $\thicksim$30x lower memory consumption than the vanilla GCN, FastGCN, and GAT.
  Besides, the proposed Bi-GCN is more effective than SGC, especially on the size of the loaded data. 

  
  For the inductive learning tasks, our binarized GNNs can significantly save the memory consumptions of both the loaded data and models, and reduce the amount of calculations with comparable performance, as shown in Table \ref{table-ind}.
  The original data size of Reddit is 534.99M, while our binarized GNNs only demand 17.61M to load the data, which proves the significance of our binarization approach.
  Note that the acceleration ratios of binarized GNNs on Reddit dataset are only $\thicksim$ 10x, because the average node degree is large, as discussed in Sec 5.3.
  In general, these results indicate that our binarization approach is efficient and it can be successfully generalized to different GNNs.

  \subsubsection{Ablation Study}
  Here, an ablation study is performed to verify the effectiveness of the binarizations applied to the parameters and node features.
  As can be observed from Table \ref{table-trans}, the prediction performances tend to vary less when the binarization is performed only to the node features.
  This phenomenon indicates that there exists many redundancies in the full-precision features and our binarization can maintain the majority portion of effective information for node classification.
  Meanwhile, the prediction results of binarizing the network parameters indicate that the binarized parameters cannot represent as much information as the full-precision parameters.
  However, if both the node attributes and parameters are binarized, a comparable performance can be achieved, compared to GCN.
  It reveals that the binarized network parameters can be effectively trained by the binarized features, i.e., Bi-GCN can successfully reduce the redundancies in the node representations, such that the useful cues can be learned well by a light-weighted binarized network.
  For the memory and computational costs, binarizing the weights and features separately will reduce the memory consumption.
  If the binarization is performed to both of them, the inference process can also be accelerated.
  
  \subsubsection{Effects of Different Model Depths}
  Figure \ref{fig4:a} shows the transductive results of GCN and Bi-GCN on Cora with different model depths.
  As can be observed, Bi-GCN is more suitable for constructing a deeper GNN than the original GCN. 
  The accuracy of GCN has dropped sharply when it consists of three or more graph convolutional layers.
  On the contrary, the performance of our Bi-GCN declines slowly.
  According to Figure \ref{fig4:b}, GCN will quickly be bothered by the overfitting issue, as the number of layers increases.
  However, our Bi-GCN can effectively alleviate this overfitting problem.
  Figure \ref{fig5} illustrates the comparisons of memory consumption and inference speed.
  Since SGC contains only one layer, its memory consumption will not change with the increase of the number of aggregations.
  When the number of layers increases, Bi-GCN can save more memories.
  For the acceleration results, the ratio between GCN and Bi-GCN tends to decrease slightly when the number of layers increases, while the actual reduced computational costs increases.
  Note that the required operations in SGC do not increase obviously because it only contains one feature extraction layer.

  \section{Conclusion}
  In this paper, we propose a binarized version of GCN, named Bi-GCN, by binarizing the network parameters and the node attributes (input data).
  The floating-point operations have been replaced by binary operations for inference acceleration.
  Besides, we design a new gradient approximation based back-propagation method to train the binarized graph convolutional layers.
  Based on our theoretical analysis, Bi-GCN can reduce the memory consumptions by an average of $\thicksim$30x for both the network parameters and node attributes, and accelerate the inference speed by an average of $\thicksim$47x, on the citation networks.
  Experiments on several datasets have demonstrated that our Bi-GCN can give a comparable performance to GCN in both the transductive and inductive tasks.
  Besides, Our binarization approach can be easily applied to other GNNs and achieve comparable results to their full-precision version.
  
  \section*{Acknowledgment}
  This work was supported in part by the National Natural Science Foundation of China under Grant 61802391, Grant U20B2069 and Grant 61972442, in part by the Natural Science Foundation of Tianjin of China under Grant 20JCYBJC00650, in part by the Natural Science Foundation of Hebei Province of China under Grant F2020202040, in part by State Key Laboratory of Software Development Environment (SKLSDE-2020ZX-18), and in part by the Fundamental Research Funds for Central Universities.

  \bibliographystyle{ieee_fullname}
  \bibliography{arxiv-v2.bib}

\end{document}